%% file: main.tex
\documentclass[10pt,twocolumn,letterpaper]{article}

\newif\ifarxiv
\arxivtrue 

\usepackage{cvpr}
\usepackage{times}
\usepackage{epsfig}
\usepackage{graphicx}
\usepackage{amsmath}
\usepackage{amssymb}
\usepackage[normalem]{ulem} 
\usepackage[T1]{fontenc} \usepackage[outline]{contour}
\usepackage{booktabs} 
\usepackage{array} 
\usepackage[thinlines]{easytable}
\usepackage{multirow}
\usepackage{footnote}
\usepackage{calc} 
\newcommand\blfootnote[1]{%
  \begingroup
  \renewcommand\thefootnote{}\footnote{#1}%
  \addtocounter{footnote}{-1}%
  \endgroup
}

\usepackage{kotex}
\usepackage{xcolor}

\contourlength{0.1pt}
\contournumber{10}

\newcommand{\arch}[1]{\textsc{#1}}
\newcommand{\stargan}{StarGAN\thinspace v2}
\newcommand{\D}{D\xspace}
\newcommand{\E}{E\xspace}
\newcommand{\F}{F\xspace}
\newcommand{\G}{G\xspace}
\newcommand{\x}{\mathbf{x}\xspace}

\newcommand{\y}{y\xspace}
\newcommand{\yt}{\widetilde{y}\xspace}

\newcommand{\z}{\mathbf{z}\xspace}
\newcommand{\s}{\mathbf{s}\xspace}
\newcommand{\st}{\widetilde{\mathbf{s}\xspace}}
\newcommand{\sh}{\hat{\mathbf{s}}\xspace}

\newcommand{\Fref}[1]{Figure \ref{#1}}
\newcommand{\Sref}[1]{Section \ref{#1}}
\newcommand{\Tref}[1]{Table \ref{#1}}
\newcommand{\Aref}[1]{Appendix \ref{#1}}

\renewcommand{\Aref}[1]{Please refer to the supplementary materials}
\include{sections/figures}
\include{sections/tables}

\usepackage{capt-of}
\usepackage[pagebackref=true,breaklinks=true,letterpaper=true,colorlinks,bookmarks=false]{hyperref}

\ifarxiv
\cvprfinalcopy
\fi

\def\cvprPaperID{20} 

\ifcvprfinal\fi
\begin{document}

\title{StarGAN v2: Diverse Image Synthesis for Multiple Domains}
\author{Yunjey Choi\textsuperscript{1}\footnotemark[1] \thinspace\quad Youngjung Uh\textsuperscript{1}\footnotemark[1] \thinspace\quad Jaejun Yoo\textsuperscript{2}\footnotemark[1] \thinspace\quad Jung-Woo Ha\textsuperscript{1} \\
\\
\textsuperscript{1}Clova AI Research, NAVER Corp. \quad  \textsuperscript{2}EPFL
}

\input{sections/0.abstract}
\input{sections/1.introduction}
\input{sections/2.method}
\input{sections/3.experiments}
\input{sections/4.discussion.tex}
\input{sections/5.related_work}
\input{sections/6.conclusion}
\input{sections/7.appendix}

{\small
\bibliographystyle{ieee}
\bibliography{citation}
}

\end{document}

%% file: sections/figures.tex
\newcommand{\figarch}{
\begin{figure*}[t]
\vspace{-5mm}
\centering
\includegraphics[width=1.0\linewidth]{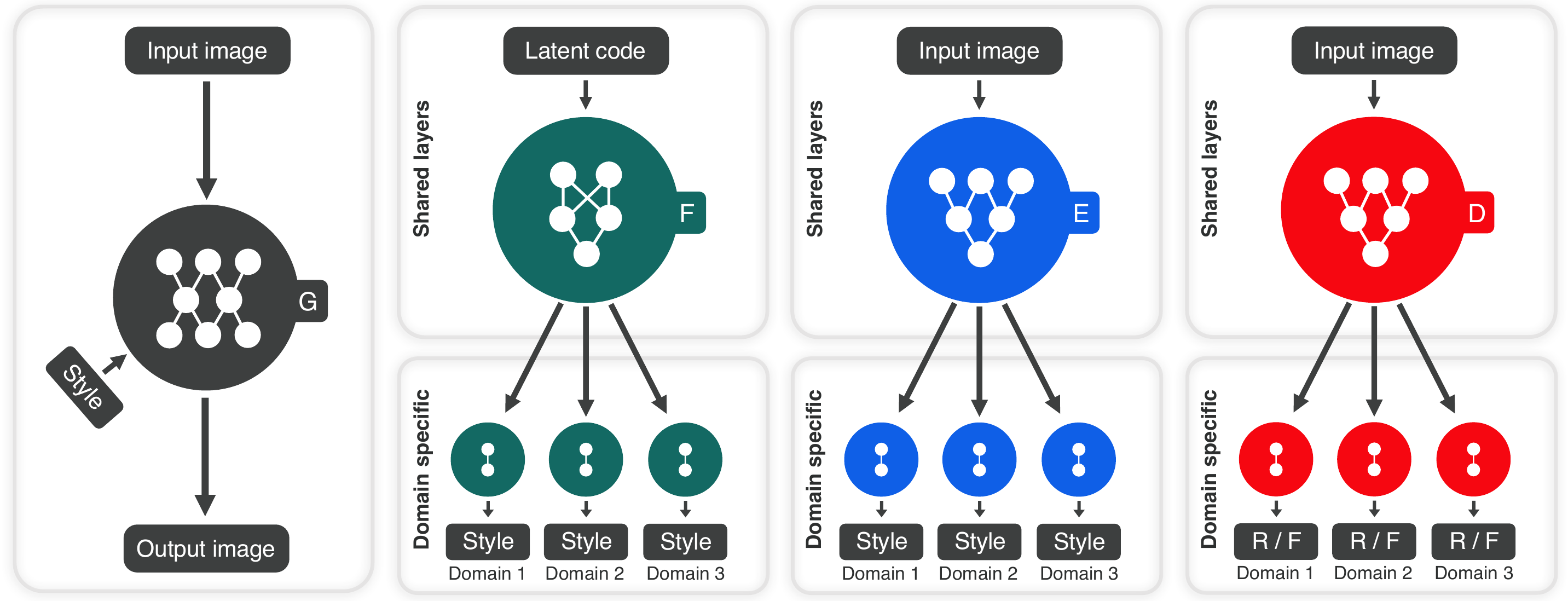}\\
\makebox[0.25\linewidth][c]{\footnotesize{(a) Generator}}\hfill
\makebox[0.25\linewidth][c]{\footnotesize{(b) Mapping network}}\hfill
\makebox[0.25\linewidth][c]{\footnotesize{(c) Style encoder}}\hfill
\makebox[0.25\linewidth][c]{\footnotesize{(d) Discriminator}}\hfill \\
\vspace{1.9mm}
\caption{Overview of \stargan, consisting of four modules.
\textbf{(a)} The generator translates an input image into an output image
reflecting the domain-specific style code. \textbf{(b)} The mapping network transforms a latent code into style codes for multiple domains, one of which is randomly selected during training. \textbf{(c)} The style encoder extracts the style code of an image, allowing the generator to perform reference-guided image synthesis. \textbf{(d)} The discriminator distinguishes between real and fake images from multiple domains. Note that all modules except the generator contain multiple output branches, one of which is selected when training the corresponding domain.} 
\label{fig:networks}
\vspace{-2mm}
\end{figure*}
}

\newcommand{\figablation}{
\renewcommand{\h}{41mm}
\begin{figure}[t]
\vspace{-2mm}
\includegraphics[width=\h]{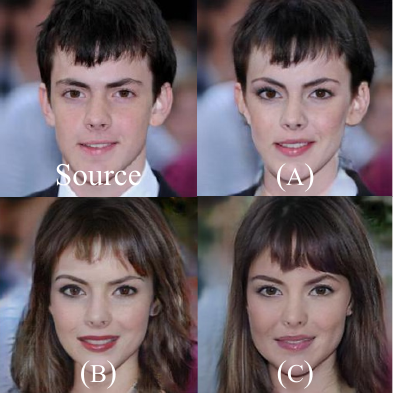}\hspace{-0.1mm}
\includegraphics[width=\h]{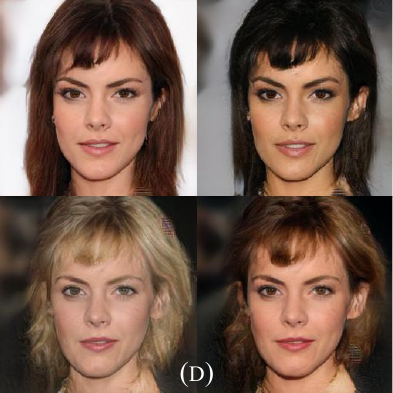}\vspace{0.4mm}\\
\includegraphics[width=\h]{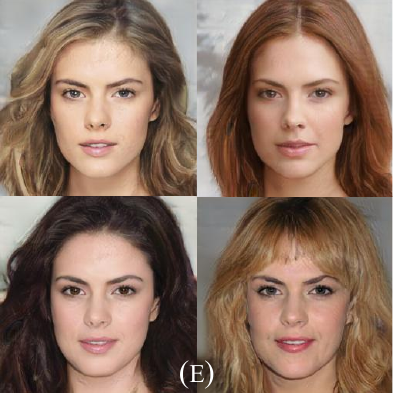}\hspace{-0.1mm}
\includegraphics[width=\h]{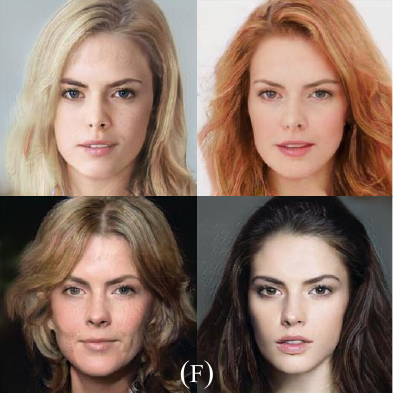}\\
\vspace{-3mm}
\caption{Visual comparison of generated images using each configuration in \Tref{tab:ablation}. Note that given a source image, the configurations (\arch{a}) - (\arch{c}) provide a single output, while (\arch{d}) - (\arch{f}) generate multiple output images.}\vspace{-3mm}
\label{fig:ablation}
\end{figure}
}

\newcommand{\figrefceleba}{
\begin{figure*}[t]
\centering
\begin{minipage}[c]{0.972\linewidth}
\includegraphics[width=1.0\linewidth]{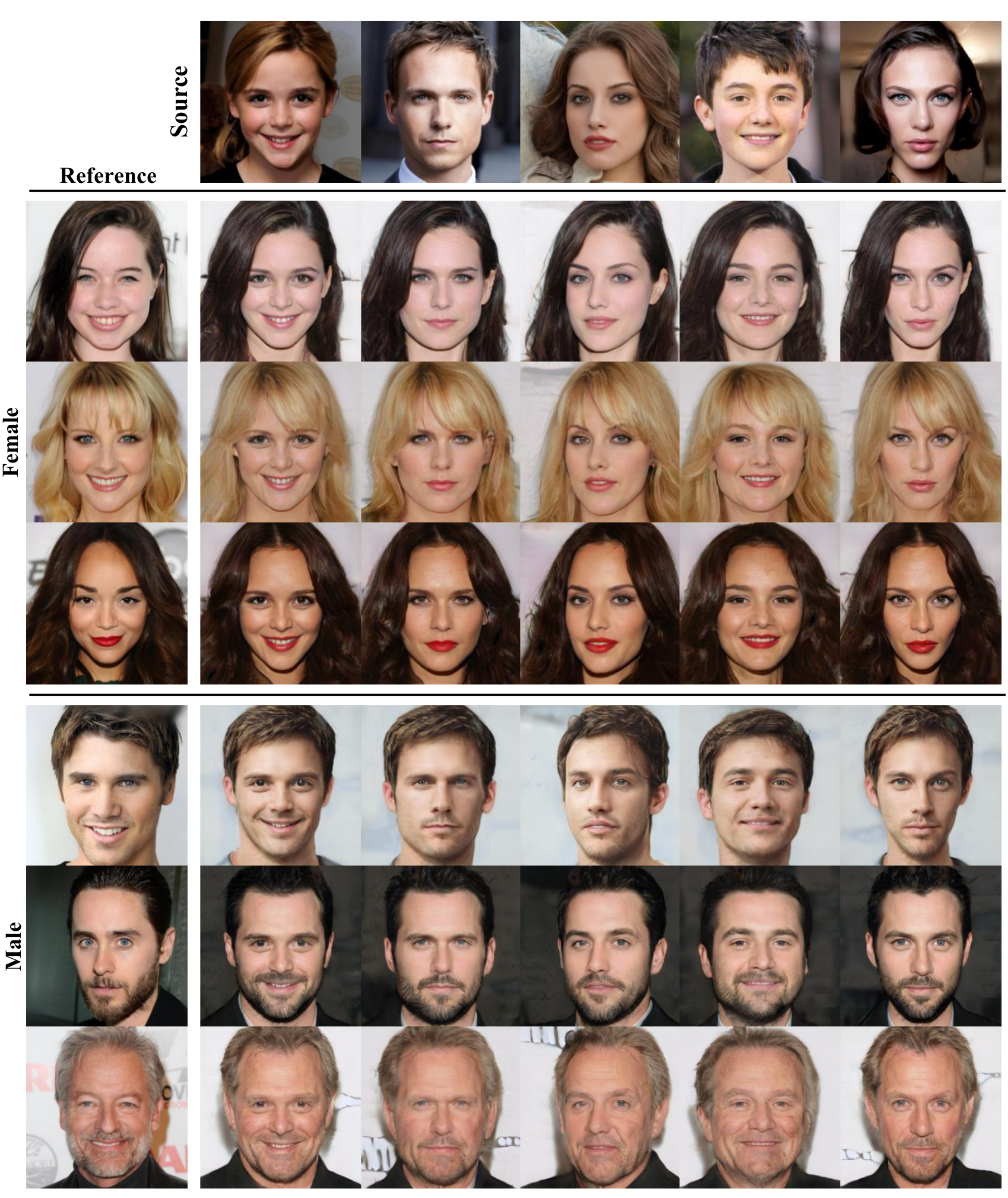}\\
\vspace{-1mm}
\end{minipage}
\caption{Reference-guided image synthesis results on CelebA-HQ. 
The source and reference images in the first row and the first column are real images, while the rest are images generated by our proposed model, \stargan. Our model learns to transform a source image reflecting the style of a given reference image. High-level semantics such as hairstyle, makeup, beard and age are followed from the reference images, while the pose and identity of the source images are preserved. Note that the images in each column share a single identity with different styles, and those in each row share a style with different identities. 
}
\label{fig:refceleba}
\vspace{-1mm}
\end{figure*}
}

\newcommand{\figrandom}{
\renewcommand{\h}{17.0mm}
\newcommand{\hdataset}{1mm}
\newcommand{\hsrc}{0.5mm}
\newcommand{\himg}{-0.6mm}
\begin{figure*}[t]
\vspace{-10mm}
\footnotesize{
\makebox[\h][c]{\textbf{Source}}\hspace{\hsrc}
\makebox[\h][c]{MUNIT~\cite{huang2018munit}}\hspace{\himg}
\makebox[\h][c]{DRIT~\cite{lee2018drit}}\hspace{\himg}
\makebox[\h][c]{MSGAN~\cite{mao2019msgan}}\hspace{\himg}
\makebox[\h][c]{\textbf{Ours}}\hspace{\hdataset}
\makebox[\h][c]{\textbf{Source}}\hspace{\hsrc}
\makebox[\h][c]{MUNIT~\cite{huang2018munit}}\hspace{\himg}
\makebox[\h][c]{DRIT~\cite{lee2018drit}}\hspace{\himg}
\makebox[\h][c]{MSGAN~\cite{mao2019msgan}}\hspace{\himg}
\makebox[\h][c]{\textbf{Ours}}\vspace{0.1mm}
}\\\
\renewcommand{\himg}{-1mm}
\includegraphics[width=\h]{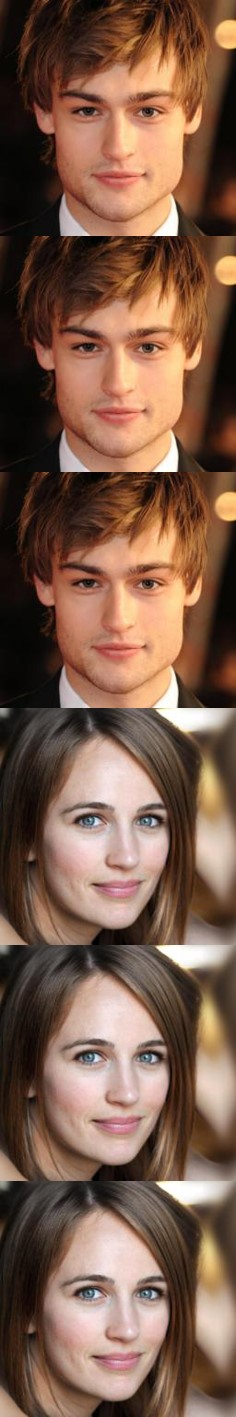}\hspace{\hsrc}
\includegraphics[width=\h]{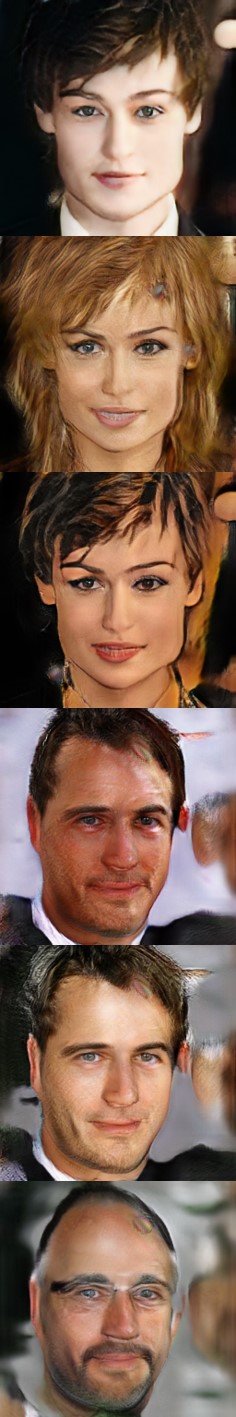}\hspace{\himg}
\includegraphics[width=\h]{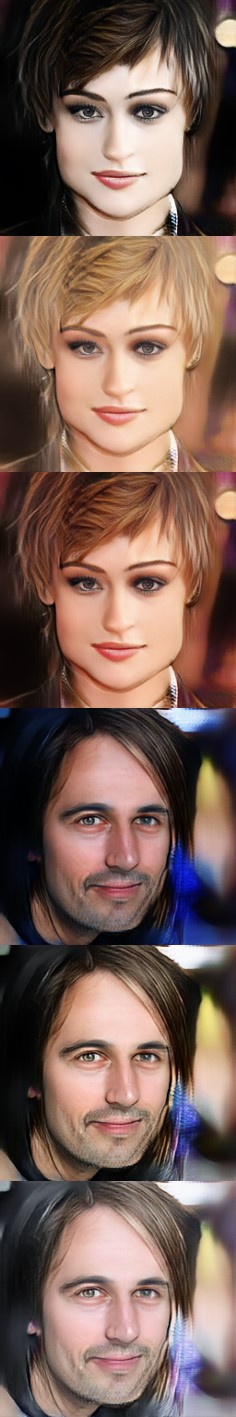}\hspace{\himg}
\includegraphics[width=\h]{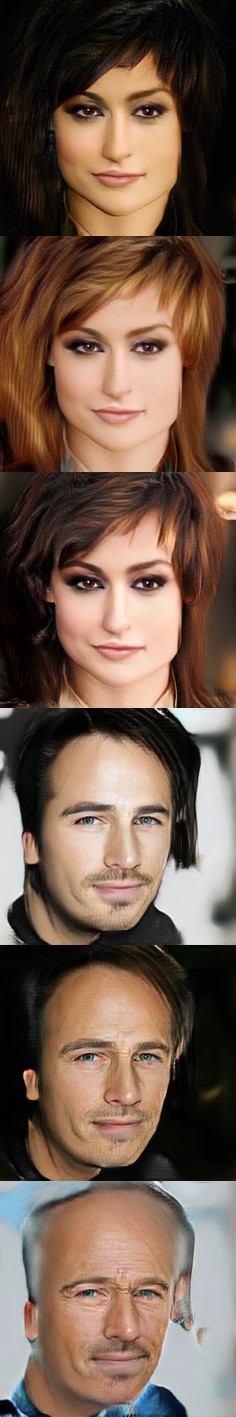}\hspace{\himg}
\includegraphics[width=\h]{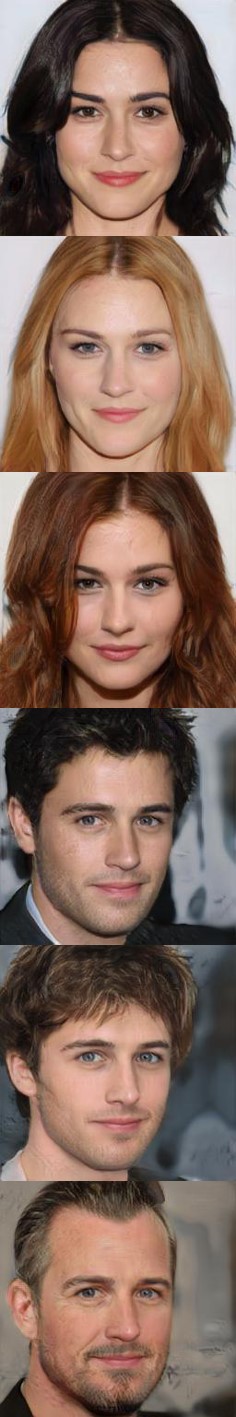}\hspace{\hdataset}
\includegraphics[width=\h]{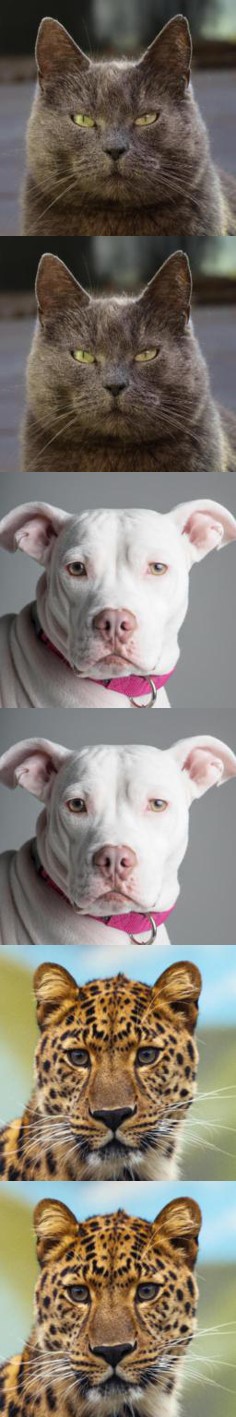}\hspace{\hsrc}
\includegraphics[width=\h]{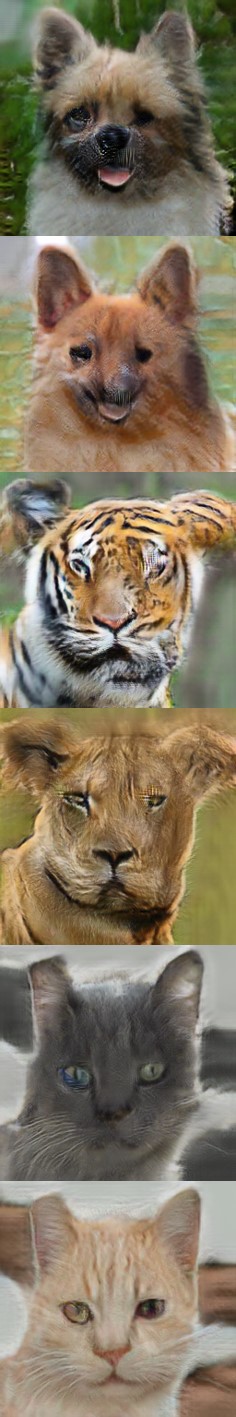}\hspace{\himg}
\includegraphics[width=\h]{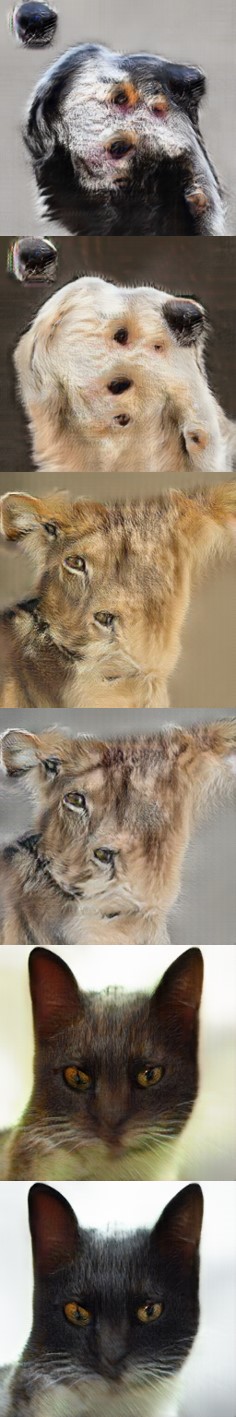}\hspace{\himg}
\includegraphics[width=\h]{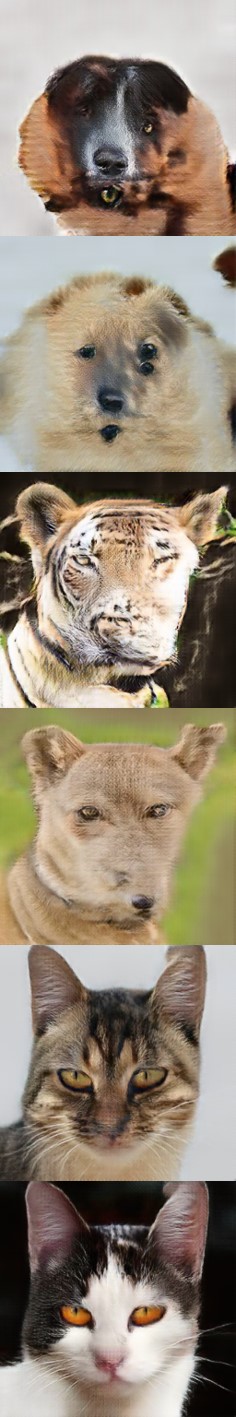}\hspace{\himg}
\includegraphics[width=\h]{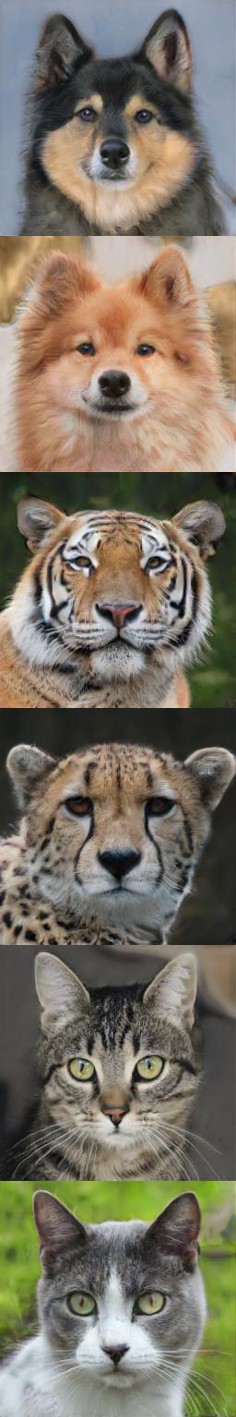}\vspace{0.5mm}
\makebox[0.5\linewidth][c]{\footnotesize{(a) Latent-guided synthesis on CelebA-HQ}}\hfill
\makebox[0.5\linewidth][c]{\footnotesize{(b) Latent-guided synthesis on AFHQ}}\hfill
\vspace{1mm}
\caption{Qualitative comparison of latent-guided image synthesis results on the CelebA-HQ and AFHQ datasets. Each method translates the source images (left-most column) to target domains using randomly sampled latent codes. \textbf{(a)} The top three rows correspond to the results of converting male to female and vice versa in the bottom three rows. \textbf{(b)} Every two rows from the top show the synthesized images in the following order: cat-to-dog, dog-to-wildlife, and wildlife-to-cat.}
\vspace{-2mm}
\label{fig:latentguided}
\end{figure*}
}

\newcommand{\figref}{
\renewcommand{\h}{17mm}
\renewcommand{\hdataset}{1mm}
\renewcommand{\href}{0.5mm}
\renewcommand{\himg}{-0.7mm}
\begin{figure*}[t]
\vspace{-3mm}
\footnotesize{
\makebox[\h][c]{\textbf{Source}}\hspace{\himg}
\makebox[\h][c]{\textbf{Reference}}\hspace{\href}
\makebox[\h][c]{DRIT~\cite{lee2018drit}}\hspace{\himg}
\makebox[\h][c]{MSGAN~\cite{mao2019msgan}}\hspace{\himg}
\makebox[\h][c]{\textbf{Ours}}\hspace{\hdataset}
\makebox[\h][c]{\textbf{Source}}\hspace{\himg}
\makebox[\h][c]{\textbf{Reference}}\hspace{\href}
\makebox[\h][c]{DRIT~\cite{lee2018drit}}\hspace{\himg}
\makebox[\h][c]{MSGAN~\cite{mao2019msgan}}\hspace{\himg}
\makebox[\h][c]{\textbf{Ours}}\vspace{0.1mm}
}\\\
\renewcommand{\himg}{-1mm}
\includegraphics[width=\h]{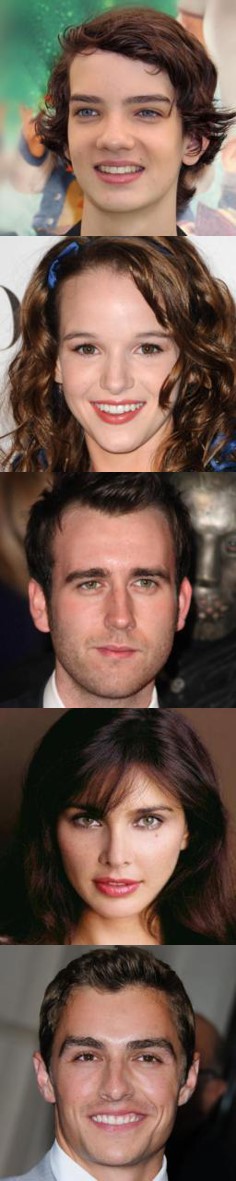}\hspace{\himg}
\includegraphics[width=\h]{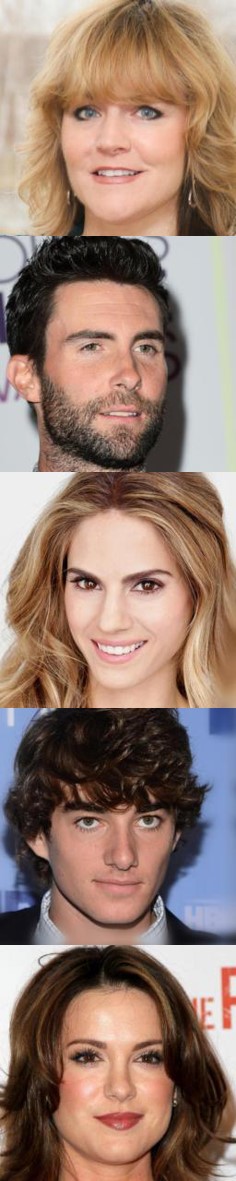}\hspace{\href}
\includegraphics[width=\h]{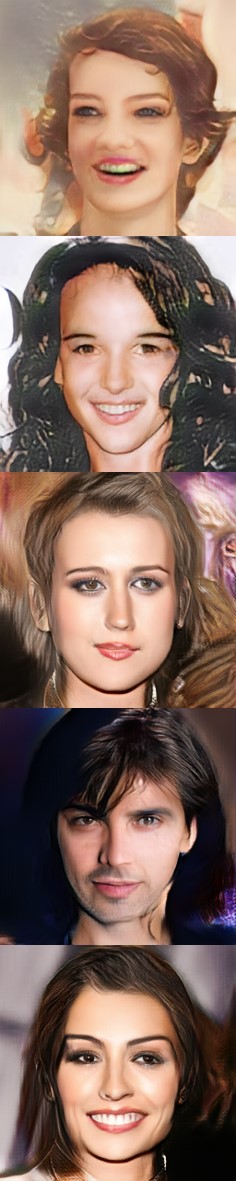}\hspace{\himg}
\includegraphics[width=\h]{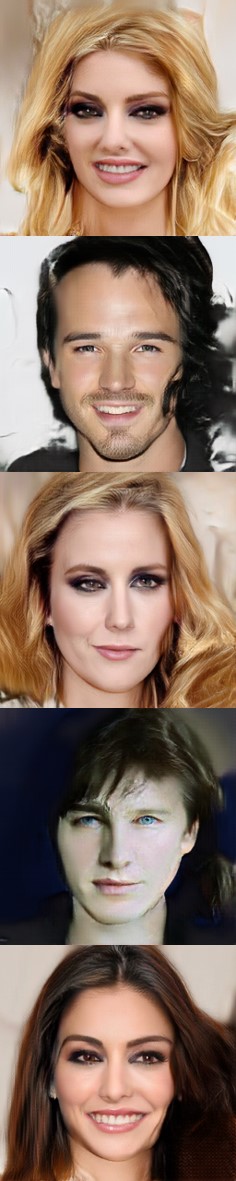}\hspace{\himg}
\includegraphics[width=\h]{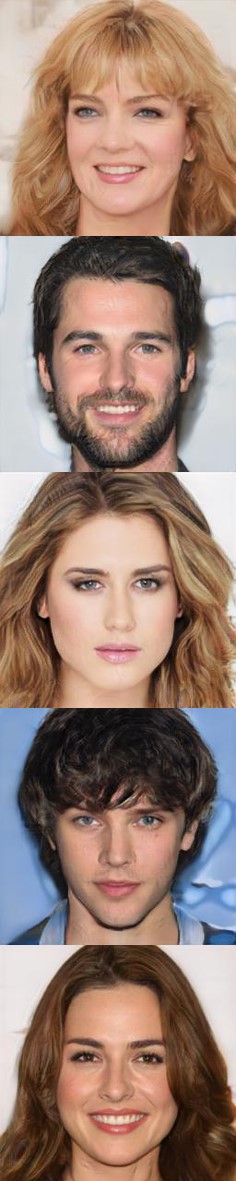}\hspace{\hdataset}
\includegraphics[width=\h]{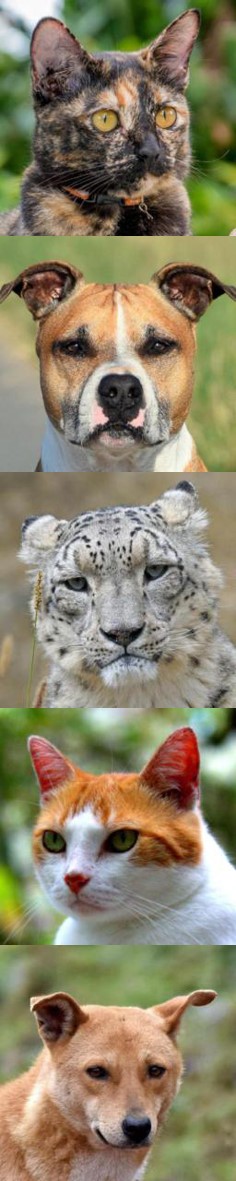}\hspace{\himg}
\includegraphics[width=\h]{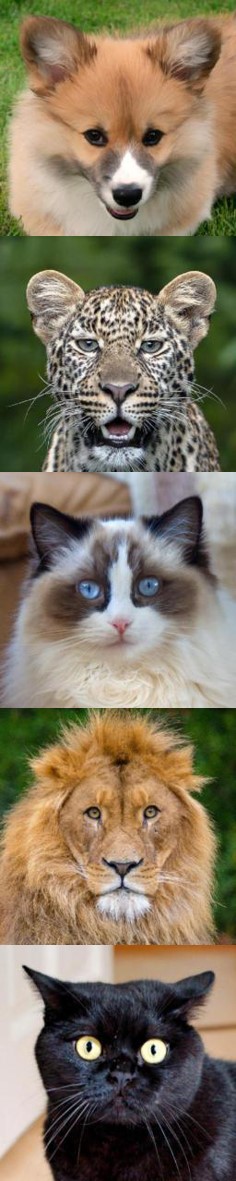}\hspace{\href}
\includegraphics[width=\h]{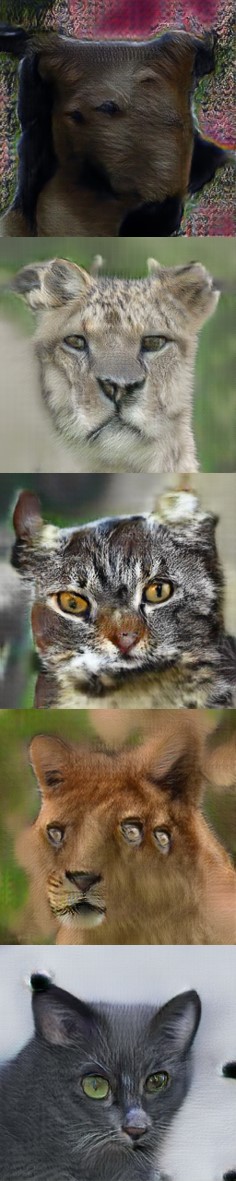}\hspace{\himg}
\includegraphics[width=\h]{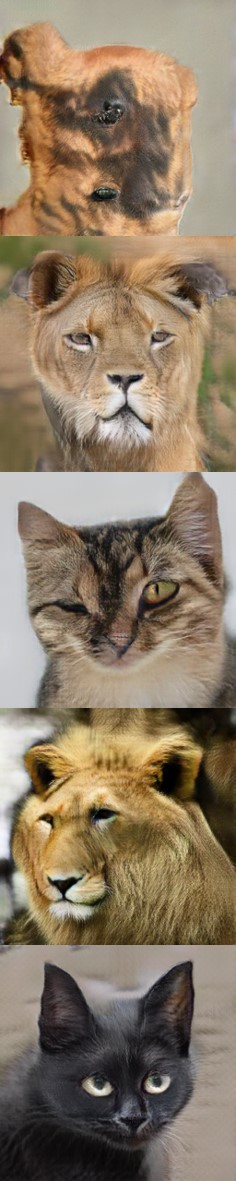}\hspace{\himg}
\includegraphics[width=\h]{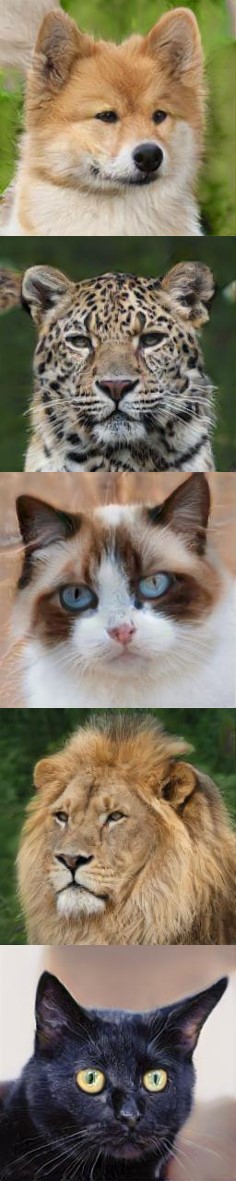}\vspace{0.5mm}
\makebox[0.5\linewidth][c]{\footnotesize{(a) Reference-guided synthesis on CelebA-HQ}}\hfill
\makebox[0.5\linewidth][c]{\footnotesize{(b) Reference-guided synthesis on AFHQ}}\hfill
\vspace{0.5mm}
\caption{Qualitative comparison of reference-guided image synthesis results on the CelebA-HQ and AFHQ datasets. Each method translates the source images into target domains, reflecting the styles of the reference images.}
\label{fig:refcompare}
\end{figure*}
}

\newcommand{\figffhq}{
\renewcommand{\h}{27.0mm}
\renewcommand{\himg}{-0mm}
\begin{figure}[t]
\vspace{-11mm}
\makebox[\h][c]{Source}\hspace{\himg}
\makebox[\h][c]{Reference}\hspace{\himg}
\makebox[\h][c]{Output}\vspace{0.3mm}
\\\
\includegraphics[width=\h]{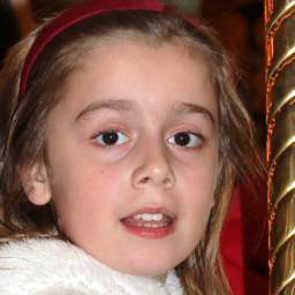}\hspace{\himg}
\includegraphics[width=\h]{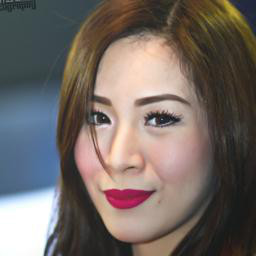}\hspace{\himg}
\includegraphics[width=\h]{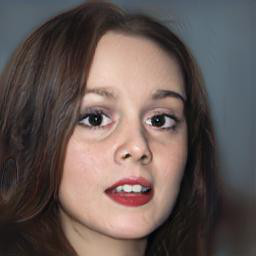}\vspace{-1mm}\\
\includegraphics[width=\h]{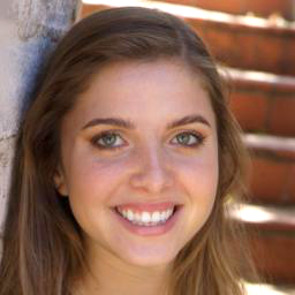}\hspace{\himg}
\includegraphics[width=\h]{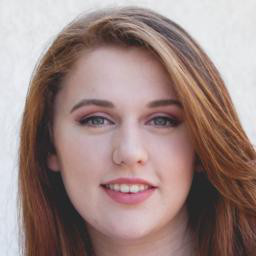}\hspace{\himg}
\includegraphics[width=\h]{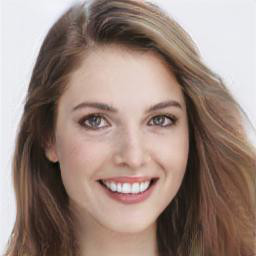}\vspace{-1mm}\\
\includegraphics[width=\h]{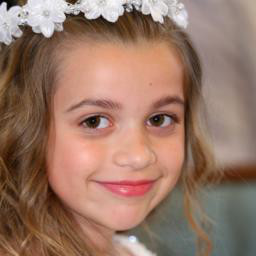}\hspace{\himg}
\includegraphics[width=\h]{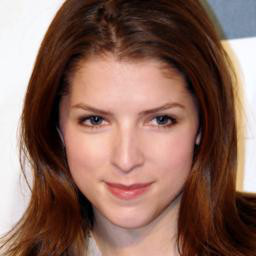}\hspace{\himg}
\includegraphics[width=\h]{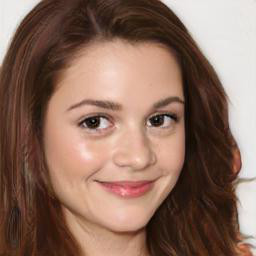}\vspace{0mm}
\caption{Reference-guided synthesis results on FFHQ with the model trained on CelebA-HQ. Despite the distribution gap between the two datasets, \stargan{} successfully extracts the style codes of the references and synthesizes faithful images.}
\label{fig:ffhq}
\vspace{-3mm}
\end{figure}
}

\newcommand{\figafhqdataset}{
\begin{figure*}[t]
\centering
\includegraphics[width=1.0\linewidth]{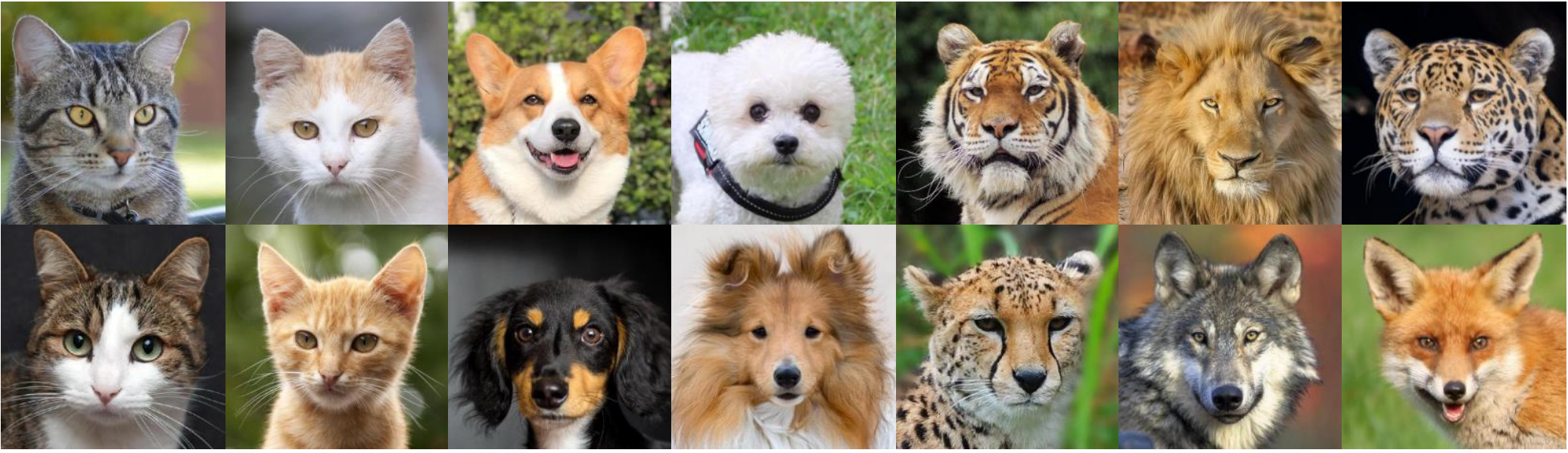}\\
\vspace{0mm}
\caption{Examples from our newly collected AFHQ dataset.}
\label{fig:afhq_dataset}
\vspace{-1mm}
\end{figure*}
}

\newcommand{\figrefcelebatwo}{
\begin{figure*}[t]
\centering
\begin{minipage}[c]{0.972\linewidth}
\includegraphics[width=1.0\linewidth]{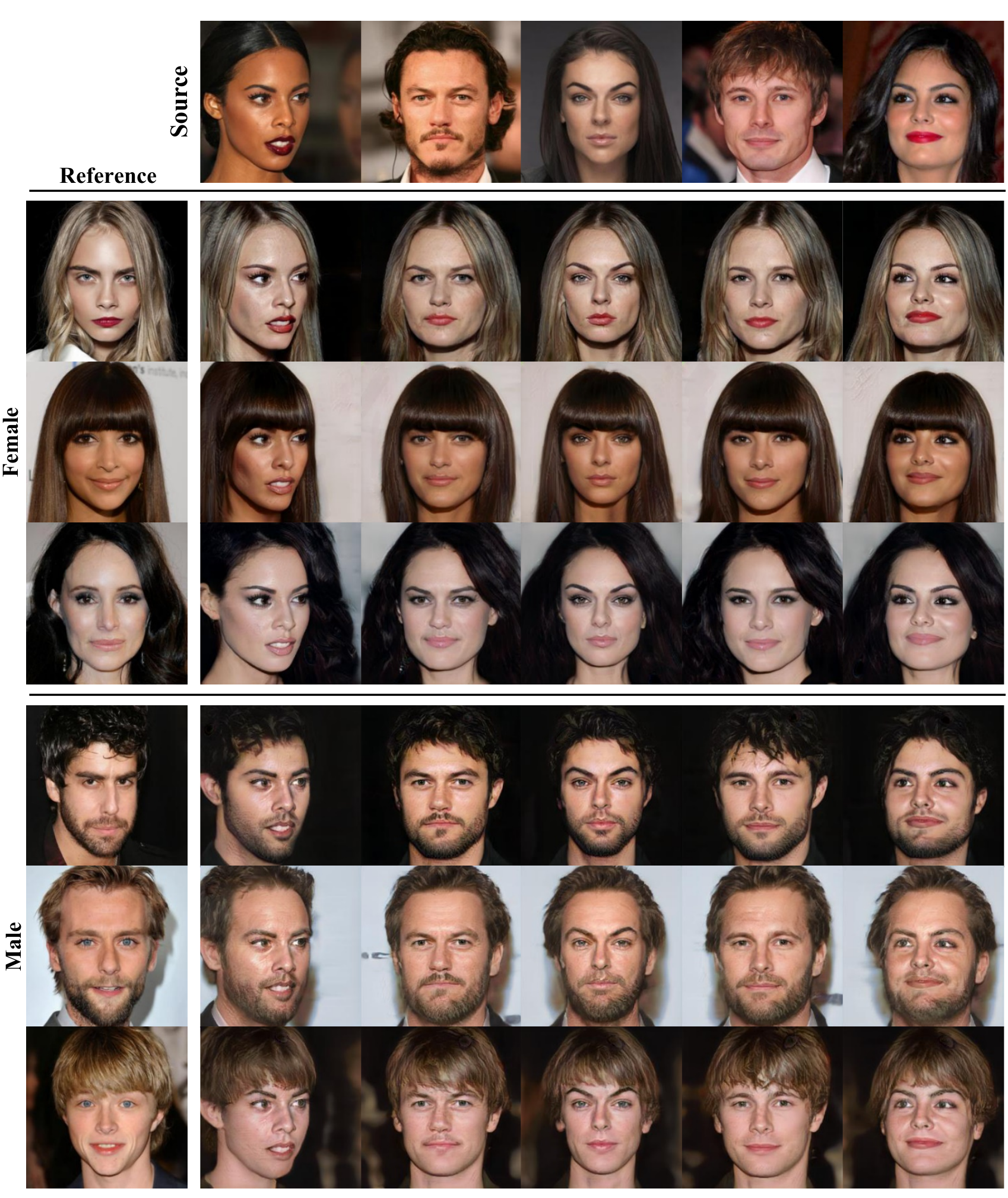}\\
\vspace{-1mm}
\end{minipage}
\caption{Reference-guided image synthesis results on CelebA-HQ. 
The source and reference images in the first row and the first column are real images, while the rest are images generated by our proposed model, \stargan. Our model learns to transform a source image reflecting the style of a given reference image. High-level semantics such as hairstyle, makeup, beard and age are followed from the reference images, while the pose and identity of the source images are preserved. Note that the images in each column share a single identity with different styles, and those in each row share a style with different identities. 
}
\label{fig:refcelebatwo}
\vspace{-1mm}
\end{figure*}
}

\newcommand{\figrefafhq}{
\begin{figure*}[t]
\vspace{-5mm}
\centering
\begin{minipage}[c]{1.0\linewidth}
\includegraphics[width=1.0\linewidth]{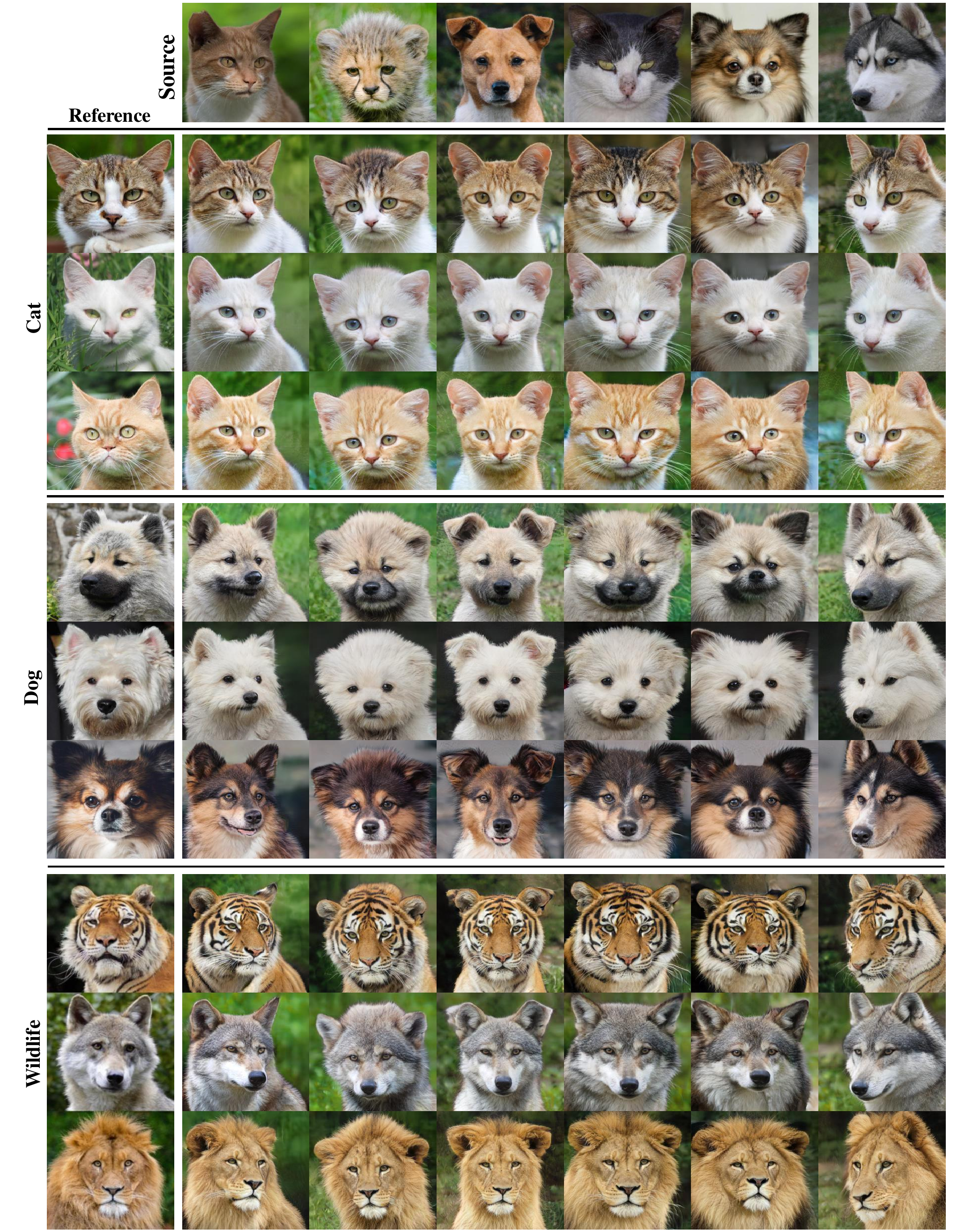}\\
\vspace{-2mm}
\end{minipage}
\caption{Reference-guided image synthesis results on AFHQ. All images except the sources and references are generated by our proposed model, \stargan{}. High-level semantics such as hair are followed from the references, while the pose of the sources are preserved.}
\label{fig:refafhq}
\vspace{-1mm}
\end{figure*}
}

%% file: sections/tables.tex
\newcommand{\h}{0mm}

\newcommand{\tabAblation}{
\begin{table}[t]
\vspace{-2mm}
\centering
\newcolumntype{x}{>{\centering\arraybackslash\hspace{0pt}}p{13.9mm}}
\footnotesize{
\begin{tabular}{|l@{\hspace{1.5mm}}l|x|x|}
\hline
  & \textbf{Method} & \textbf{FID} & \textbf{LPIPS} \\\hline\hline
\arch{a} & Baseline StarGAN \cite{choi2018stargan} \ \   & 98.4           & -          \\\hline %
\arch{b} & + Multi-task discriminator       & 91.4           & -          \\\hline %
\arch{c} & + Tuning (\eg ${R}_{1}$ regularization)       & 80.5           & -          \\\hline %
\arch{d} & + Latent code injection    & 32.3          & 0.312          \\\hline %
\arch{e} & + Replace (\arch{d}) with style code     & 17.1           & 0.405          \\\hline %
\arch{f} & + Diversity regularization      & \textbf{13.7}   & \textbf{0.452}          \\\hline %
\end{tabular}
}
\vspace{1mm}
\caption{Performance of various configurations on CelebA-HQ. Frechet inception distance (FID) indicates the distance between two distributions of real and generated images (lower is better), while learned perceptual image patch similarity (LPIPS) measures the diversity of generated images (higher is better).} 
\label{tab:ablation}
\vspace{-1mm}
\end{table}
}

\newcommand{\tabRandom}{
\begin{table}[t]
\centering
\newcolumntype{x}{>{\centering\arraybackslash\hspace{0pt}}p{11.2mm}}
\footnotesize{
\begin{tabular}{|l@{\hspace{1.5mm}}l|x|x|x|x|}
\hline
&
& \multicolumn{2}{c|}{\raisebox{0mm}[2mm][1mm]{\textbf{CelebA-HQ}}}
& \multicolumn{2}{c|}{\raisebox{0mm}[2mm][1mm]{\textbf{AFHQ}}}
\\
\cline{3-6}
  \multicolumn{2}{|l|}{\textbf{\space Method}}
  & \multicolumn{1}{c|}{\raisebox{-0.2mm}[2mm][0mm]{\textbf{FID}}}
  & \multicolumn{1}{c|}{\raisebox{-0.2mm}[2mm][0mm]{\textbf{LPIPS}}}
  & \multicolumn{1}{c|}{\raisebox{-0.2mm}[2mm][0mm]{\textbf{FID}}}
  & \multicolumn{1}{c|}{\raisebox{-0.2mm}[2mm][0mm]{\textbf{LPIPS}}} \\\hline\hline
& MUNIT \cite{huang2018munit} \ \  & \raisebox{0mm}[2.5mm][0mm]{31.4}           & 0.363            & 41.5    & 0.511 \\ %
& DRIT \cite{lee2018drit}       & 52.1           & 0.178            & 95.6    & 0.326 \\ %
& MSGAN \cite{mao2019msgan}      & 33.1           & 0.389            & 61.4    & \textbf{0.517} \\ %
& \stargan  & \textbf{13.7}      & \textbf{0.452}   & \textbf{16.2}    & 0.450 \\ %
\hline\hline
& Real images  & \raisebox{0mm}[2.5mm][0mm]{14.8}           & -            & 12.9    & - \\\hline %
\end{tabular}
}\vspace{0.25mm}
\caption{Quantitative comparison on latent-guided synthesis. 
The FIDs of real images are computed between the training and test sets. Note that they may not be optimal values since the number of test images is insufficient, but we report them for reference.}
\label{tab:random}
\vspace{-3mm}
\end{table}
}

\newcommand{\tabReference}{
\begin{table}[t]
\centering
\newcolumntype{x}{>{\centering\arraybackslash\hspace{0pt}}p{11.2mm}}
\footnotesize{
\begin{tabular}{|l@{\hspace{1.5mm}}l|x|x|x|x|}
\hline
&
& \multicolumn{2}{c|}{\raisebox{-0.1mm}[2mm][1mm]{\textbf{CelebA-HQ}}}
& \multicolumn{2}{c|}{\raisebox{-0.1mm}[2mm][1mm]{\textbf{AFHQ}}}
\\
\cline{3-6}
  \multicolumn{2}{|l|}{\textbf{\space Method}}
  & \multicolumn{1}{c|}{\raisebox{-0.2mm}[2mm][0mm]{\textbf{FID}}}
  & \multicolumn{1}{c|}{\raisebox{-0.2mm}[2mm][0mm]{\textbf{LPIPS}}}
  & \multicolumn{1}{c|}{\raisebox{-0.2mm}[2mm][0mm]{\textbf{FID}}}
  & \multicolumn{1}{c|}{\raisebox{-0.2mm}[2mm][0mm]{\textbf{LPIPS}}} \\\hline\hline
& MUNIT \cite{huang2018munit}  & \raisebox{0mm}[2.5mm][0mm]{107.1}           &    0.176        & 223.9    & 0.199 \\ %
& DRIT \cite{lee2018drit}       & 53.3           & 0.311            & 114.8    & 0.156 \\ 
& MSGAN \cite{mao2019msgan}     & 39.6           & 0.312            & 69.8    & 0.375 \\ 
& \stargan  & \textbf{23.8}     & \textbf{0.388}      & \textbf{19.8}    & \textbf{0.432} \\ 
\hline\hline
& Real images  & \raisebox{0mm}[2.5mm][0mm]{14.8}           & -            & 12.9    & - \\\hline 
\end{tabular}
}\vspace{0.25mm}
\caption{Quantitative comparison on reference-guided synthesis. We sample ten reference images to synthesize diverse images.
} 
\label{tab:reference}
\vspace{-3mm}
\end{table}
}

\newcommand{\tabUser}{
\begin{table}[t]
\vspace{-7mm}
\centering
\newcolumntype{x}{>{\centering\arraybackslash\hspace{0pt}}p{11mm}}
\footnotesize{
\begin{tabular}{|l@{\hspace{1.5mm}}l|x|x|x|x|}
\hline
&
& \multicolumn{2}{c|}{\raisebox{-0.1mm}[2mm][1mm]{\textbf{CelebA-HQ}}}
& \multicolumn{2}{c|}{\raisebox{-0.1mm}[2mm][1mm]{\textbf{AFHQ}}}
\\
\cline{3-6}
  \multicolumn{2}{|l|}{\textbf{\space Method}}
  & \multicolumn{1}{c|}{\raisebox{-0.2mm}[2mm][0mm]{\textbf{Quality}}}
  & \multicolumn{1}{c|}{\raisebox{-0.2mm}[2mm][0mm]{\textbf{Style}}}
  & \multicolumn{1}{c|}{\raisebox{-0.2mm}[2mm][0mm]{\textbf{Quality}}}
  & \multicolumn{1}{c|}{\raisebox{-0.2mm}[2mm][0mm]{\textbf{Style}}} \\\hline\hline
& MUNIT \cite{huang2018munit}    & \raisebox{0mm}[2.5mm][0mm]{6.2}  & 7.4   & 1.6   & 0.2 \\
& DRIT \cite{lee2018drit}        & 11.4  & 7.6   & 4.1   & 2.8 \\
& MSGAN \cite{mao2019msgan}      & 13.5  & 10.1   & 6.2   & 4.9 \\
& \stargan  & \textbf{68.9}     & \textbf{74.8}      & \textbf{88.1}    & \textbf{92.1} \\
\hline
\end{tabular}
}
\caption{Votes from AMT workers for the most preferred method regarding visual quality and style reflection (\%). \stargan{} outperforms the baselines with remarkable margins in all aspects.
}
\label{tab:userstudy}
\vspace{-5mm}
\end{table}
}

\newcommand{\shape}[6]{$\makebox[\widthof{#1}][c]{#4}\times\makebox[\widthof{#2}][c]{#5}\times\makebox[\widthof{#3}][c]{#6}$}

\newcommand{\networkG}{
\begin{table}[h]
\vspace{-3mm}
\centering
\begin{tabular}{lccc}
\toprule
\textsc{Layer} & \textsc{Resample} & \textsc{Norm} & \textsc{Output Shape}\vspace{0.5mm}\\
\toprule 
Image $\x$ & - & - & \shape{256}{256}{512}{256}{256}{3}\\
\midrule
Conv1$\times$1 & - & - & \shape{256}{256}{16ch}{256}{256}{64}\\
ResBlk & AvgPool & IN & \shape{256}{256}{16ch}{128}{128}{128}\\
ResBlk & AvgPool & IN & \shape{256}{256}{16ch}{64}{64}{256}\\
ResBlk & AvgPool & IN & \shape{256}{256}{16ch}{32}{32}{512}\\
ResBlk & AvgPool & IN & \shape{256}{256}{16ch}{16}{16}{512}\\
\midrule
ResBlk & - & IN & \shape{256}{256}{16ch}{16}{16}{512}\\
ResBlk & - & IN & \shape{256}{256}{16ch}{16}{16}{512}\\
ResBlk & - & AdaIN & \shape{256}{256}{16ch}{16}{16}{512}\\
ResBlk & - & AdaIN & \shape{256}{256}{16ch}{16}{16}{512}\\
\midrule
ResBlk & Upsample & AdaIN & \shape{256}{256}{512}{32}{32}{512}\\
ResBlk & Upsample & AdaIN & \shape{256}{256}{512}{64}{64}{256}\\
ResBlk & Upsample & AdaIN & \shape{256}{256}{512}{128}{128}{128}\\
ResBlk & Upsample & AdaIN & \shape{256}{256}{512}{256}{256}{64}\\
Conv1$\times$1 & - & - & \shape{256}{256}{512}{256}{256}{3}\\
\bottomrule
\end{tabular}\vspace{2mm}
\caption{Generator architecture.}
\label{tab:generator_architecture}
\end{table}
}

\newcommand{\networkF}{
\begin{table}[h]
\centering
\begin{tabular}{lccc}
\toprule
\textsc{Type} & \textsc{Layer} & \textsc{Actvation} & \textsc{Output shape}\vspace{0.5mm}\\
\toprule 
Shared & Latent $\z$ & - & 16\\
\midrule 
Shared & Linear & ReLU & 512\\
Shared & Linear & ReLU & 512\\
Shared & Linear & ReLU & 512\\
Shared & Linear & ReLU & 512\\
\midrule
Unshared & Linear & ReLU & 512\\
Unshared & Linear & ReLU & 512\\
Unshared & Linear & ReLU & 512\\
Unshared & Linear & - & 64 \\
\bottomrule
\end{tabular}\vspace{2mm}
\caption{Mapping network architecture.}
\label{tab:mappingnetwork_architecture}
\end{table}
}

\newcommand{\networkE}{
\begin{table}[h]
\centering
\begin{tabular}{lccc}
\toprule
\textsc{Layer} & \textsc{Resample} & \textsc{Norm} & \textsc{Output shape}\vspace{0.5mm}\\
\toprule 
Image $\x$ & - & - & \shape{256}{256}{32ch}{256}{256}{3}\\
\midrule
Conv1$\times$1 & - & - & \shape{256}{256}{16ch}{256}{256}{64}\\
ResBlk & AvgPool & - & \shape{256}{256}{32ch}{128}{128}{128}\\
ResBlk & AvgPool & - & \shape{256}{256}{32ch}{64}{64}{256}\\
ResBlk & AvgPool & - & \shape{256}{256}{32ch}{32}{32}{512}\\
ResBlk & AvgPool & - & \shape{256}{256}{32ch}{16}{16}{512}\\
ResBlk & AvgPool & - & \shape{256}{256}{32ch}{8}{8}{512}\\
ResBlk & AvgPool & - & \shape{256}{256}{32ch}{4}{4}{512}\\
\midrule
LReLU & - & - & \shape{256}{256}{32ch}{4}{4}{512}\\
Conv4$\times$4 & - & - & \shape{256}{256}{32ch}{1}{1}{512}\\
LReLU & - & - & \shape{256}{256}{32ch}{1}{1}{512}\\
\midrule
Reshape & - & - & 512 \\
Linear \texttt{*} \texttt{K} & - & - & \texttt{D} \texttt{*} \texttt{K} \\
\bottomrule
\end{tabular}\vspace{2mm}
\caption{Style encoder and discriminator architectures. \texttt{D} and \texttt{K} represent the output dimension and number of domains, respectively.}
\label{tab:encoder_architecture}\vspace{-7mm}
\end{table}
}

%% file: sections/0.abstract.tex
\twocolumn[{
\renewcommand\twocolumn[1][]{#1}
\maketitle
\ifarxiv
\vspace{-0.7 cm}
\fi
\begin{center}
    \centering 
    \vspace{-0.0 cm}
    \includegraphics[width=1\textwidth]{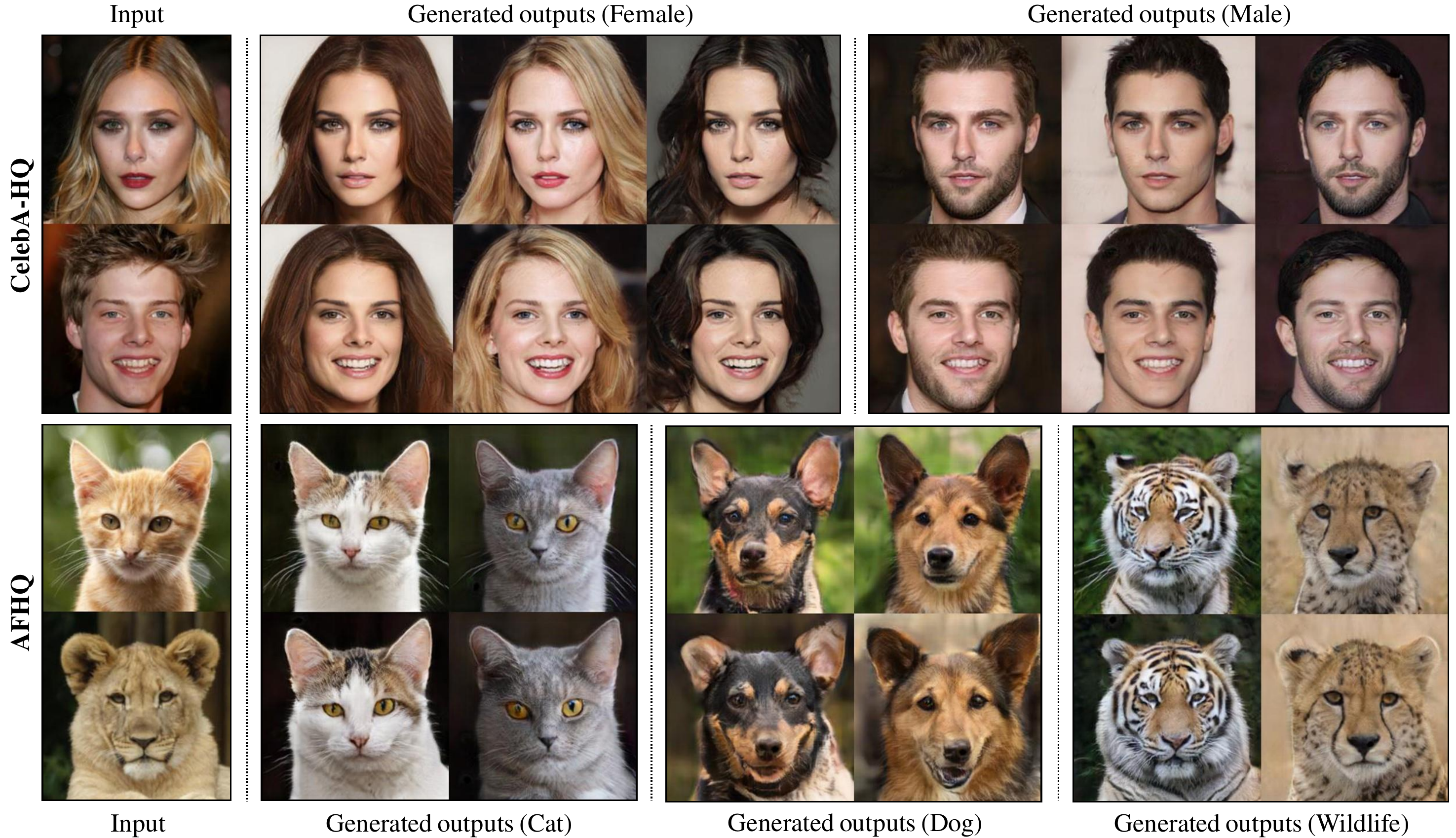}
    \captionof{figure}{Diverse image synthesis results on the CelebA-HQ dataset and the newly collected animal faces (AFHQ) dataset. The first column shows input images while the remaining columns are images synthesized by \stargan.} 
    \label{fig:teaser}
\end{center}
}]

\begin{abstract}
\vspace{-0.3cm}
A good image-to-image translation model should learn a mapping between different visual domains while satisfying the following properties: 1) diversity of generated images and 2) scalability over multiple domains. Existing methods address either of the issues, having limited diversity or multiple models for all domains. We propose \stargan{}, a single framework that tackles both and shows significantly improved results over the baselines. Experiments on CelebA-HQ and a new animal faces dataset (AFHQ) validate our superiority in terms of visual quality, diversity, and scalability. To better assess image-to-image translation models, we release AFHQ, high-quality animal faces with large inter- and intra-domain differences. The code, pretrained models, and dataset 
\ifarxiv
are available at \href{https://github.com/clovaai/stargan-v2}{clovaai/stargan-v2}. 
\fi
\end{abstract}
\vspace{-0.0 cm}

%% file: sections/1.introduction.tex
\vspace{-0.5 cm}
\section{Introduction}
\ifarxiv
\blfootnote{* indicates equal contribution}
\fi
Image-to-image translation aims to learn a mapping between different visual domains~\cite{isola2017pix2pix}. Here, \texttt{domain} implies a set of images that can be grouped as a visually distinctive category, and each image has a unique appearance, which we call \texttt{style}. For example, we can set image domains based on the gender of a person, in which case the style includes makeup, beard, and hairstyle (top half of \Fref{fig:teaser}). An ideal image-to-image translation method should be able to synthesize images considering the diverse styles in each domain. However, designing and learning such models become complicated as there can be arbitrarily large number of styles and domains in the dataset. 

\smallskip

To address the style diversity, much work on image-to-image translation has been developed~\cite{almahairi2018augcyclegan,huang2018munit,mao2019msgan,lee2018drit,na2019miso,zhu2017bicyclegan}. These methods inject a low-dimensional latent code to the generator, which can be randomly sampled from the standard Gaussian distribution. Their domain-specific decoders interpret the latent codes as recipes for various styles when generating images. However, because these methods have only considered a mapping between two domains, they are not scalable to the increasing number of domains. For example, having \texttt{K} domains, these methods require to train \texttt{K(K-1)} generators to handle translations between each and every domain, limiting their practical usage.

\smallskip

To address the scalability, several studies have proposed a unified framework~\cite{anoosheh2018combogan,choi2018stargan,hui2018domainspecificencdec,liu2019funit}. StarGAN~\cite{choi2018stargan} is one of the earliest models, which learns the mappings between all available domains using a single generator. The generator takes a domain label as an additional input, and learns to transform an image into the corresponding domain. However, StarGAN still learns a deterministic mapping per each domain, which does not capture the multi-modal nature of the data distribution.  This limitation comes from the fact that each domain is indicated by a predetermined label. Note that the generator receives a fixed label (\eg one-hot vector) as input, and thus it inevitably produces the same output per each domain, given a source image.

\smallskip

To get the best of both worlds, we propose \stargan, a scalable approach that can generate \textbf{diverse} images across \textbf{multiple domains}. In particular, we start from StarGAN and replace its domain label with our proposed domain-specific style code that can represent diverse \texttt{styles} of a specific \texttt{domain}. To this end, we introduce two modules, a mapping network and a style encoder. The mapping network learns to transform random Gaussian noise into a style code, while the encoder learns to extract the style code from a given reference image. Considering multiple domains, both modules have multiple output branches, each of which provides style codes for a specific domain. Finally, utilizing these style codes, our generator learns to successfully synthesize diverse images over multiple domains (\Fref{fig:teaser}).

\smallskip

We first investigate the effect of individual components of \stargan{} and show that our model indeed benefits from using the style code (\Sref{sec:components}). We empirically demonstrate that our proposed method is scalable to multiple domains and gives significantly better results in terms of visual quality and diversity compared to the leading methods (\Sref{sec:comparison}). Last but not least, we present a new dataset of animal faces (AFHQ) with high quality and wide variations 
\ifarxiv
(Appendix~\ref{sec:AFHQ})
\fi 
to better evaluate the performance of image-to-image translation models on large inter- and intra-domain differences. We release this dataset publicly available for research community.

%% file: sections/2.method.tex
\figarch

\section{\stargan} 
In this section, we describe our proposed framework and its training objective functions. 

\subsection{Proposed framework}
\label{sec:model_overview}
Let ${\cal X}$ and ${\cal Y}$ be the sets of images and possible domains, respectively. Given an image $\x \in {\cal X}$ and an arbitrary domain $\y \in {\cal Y}$, our goal is to train a \textbf{single} generator $\G$ that can generate \textbf{diverse} images of each domain $\y$ that corresponds to the image $\x$. We generate \textbf{domain-specific} style vectors in the learned style space of each domain and train $\G$ to reflect the style vectors. \Fref{fig:networks} illustrates an overview of our framework, which consists of four modules described below.

\medskip

\noindent\textbf{Generator} \textbf{(}\Fref{fig:networks}a\textbf{).} Our generator $\G$ translates an input image $\x$ into an output image $\G(\x, \s)$ reflecting a domain-specific style code $\s$, which is provided either by the mapping network $\F$ or by the style encoder $\E$. We use adaptive instance normalization (AdaIN)~\cite{huang2017adain,karras2019stylegan} to inject $\s$ into $G$. We observe that $\s$ is designed to represent a style of a specific domain $\y$, which removes the necessity of providing $\y$ to $\G$ and allows $G$ to synthesize images of all domains. 

\medskip

\noindent\textbf{Mapping network} \textbf{(}\Fref{fig:networks}b\textbf{).} Given a latent code $\z$ and a domain $\y$, our mapping network $\F$ generates a style code $\s = {F}_{\y}(\z)$, where ${F}_{\y}(\cdot)$ denotes an output of $\F$ corresponding to the domain $\y$. $\F$ consists of an MLP with multiple output branches to provide style codes for all available domains. $F$ can produce diverse style codes by sampling the latent vector $\z \in \cal{Z}$ and the domain $\y \in \cal{Y}$ randomly. Our multi-task architecture allows $\F$ to efficiently and effectively learn style representations of all domains.

\medskip

\noindent\textbf{Style encoder} \textbf{(}\Fref{fig:networks}c\textbf{).} Given an image $\x$ and its corresponding domain $\y$, our encoder $\E$ extracts the style code $\s = {\E}_{\y}(\x)$ of $\x$. Here, ${E}_{\y}(\cdot)$ denotes the output of $\E$ corresponding to the domain $\y$. Similar to $\F$, our style encoder $\E$ benefits from the multi-task learning setup. $\E$ can produce diverse style codes using different reference images. This allows $G$ to synthesize an output image reflecting the style $\s$ of a reference image $\x$.

\medskip

\noindent\textbf{Discriminator} \textbf{(}\Fref{fig:networks}d\textbf{).} Our discriminator $\D$ is a multi-task discriminator~\cite{liu2019funit,mescheder2018r1reg}, which consists of multiple output branches. Each branch ${D}_{y}$ learns a binary classification determining whether an image $\x$ is a real image of its domain $\y$ or a fake image $G(\x, \s)$ produced by $G$. 

\subsection{Training objectives}
\label{sec:objective_functions}
Given an image $\x\in\cal{X}$ and its original domain $y\in\cal{Y}$, we train our framework using the following objectives. 

\smallskip

\noindent \textbf{Adversarial objective.} During training, we sample a latent code $\z \in \cal{Z}$ and a target domain $\yt \in \cal{Y}$ randomly, and generate a target style code \thinspace $\st = {F}_{\yt}(\z)$. The generator $\G$ takes an image $\x$ and $\st$ as inputs and learns to generate an output image $G(\x, \st)$ via an adversarial loss
\begin{equation}
\begin{split}
 \mathcal{L}_{adv} = & \thinspace \mathbb{E}_{\x, \y} \left[ \log{{D}_{\y}(\x)} \right]  \> \>  +   \\
 & \thinspace \mathbb{E}_{\x, \yt, \z}[\log{(1 - {D}_{\yt}(G(\x, \st)))}],
\end{split}
\label{eqn:adv_loss}
\end{equation}

\noindent where ${D}_{\y}(\cdot)$ denotes the output of $\D$ corresponding to the domain $\y$. The mapping network $F$ learns to provide the style code $\st$ that is likely in the target domain $\yt$, and $G$ learns to utilize $\st$ and generate an image $G(\x, \st)$ that is indistinguishable from real images of the domain $\yt$. 
\smallskip

\noindent \textbf{Style reconstruction.} In order to enforce the generator $\G$ to utilize the style code $\st$ when generating the image $G(\x, \st)$, we employ a style reconstruction loss 

\begin{equation}
\mathcal{L}_{sty} = \mathbb{E}_{\x, \yt, \z} \left[{||\st - {E}_{\yt}(G(\x,\st))||}_{1} \right].
\label{eqn:sty_rec_loss}
\end{equation}
\noindent This objective is similar to the previous approaches \cite{huang2018munit,zhu2017bicyclegan}, which employ multiple encoders to learn a mapping from an image to its latent code. The notable difference is that we train a \textbf{single} encoder $E$ to encourage diverse outputs for multiple domains. At test time, our learned encoder $\E$ allows $\G$ to transform an input image, reflecting the style of a reference image.

\smallskip

\noindent\textbf{Style diversification.} To further enable the generator $\G$ to produce diverse images, we explicitly regularize $\G$ with the diversity sensitive loss~\cite{mao2019msgan,yang2019dsgan}
\begin{equation}
\mathcal{L}_{ds} = \mathbb{E}_{\x, \yt, \z_1, \z_2} \left[ { \lVert G(\x,\st_1) - G(\x,\st_2) \lVert}_{1} \right], 
\label{eqn:ds_loss}
\end{equation}

\noindent where the target style codes $\st_1$ and $\st_2$ are produced by $F$ conditioned on two random latent codes $\z_1$ and $\z_2$ (\ie $\st_i = {F}_{\yt}(\z_i)$ for $ i \in \{1,2\}$). Maximizing the regularization term forces $\G$ to explore the image space and  discover meaningful style features to generate diverse images. Note that in the original form, the small difference of ${\lVert \z_1 - \z_2 \lVert}_{1}$ in the denominator increases the loss significantly, which makes the training unstable due to large gradients. Thus, we remove the denominator part and devise a new equation for stable training but with the same intuition.

\smallskip 

\noindent\textbf{Preserving source characteristics.} To guarantee that the generated image $G(\x, \st)$ properly preserves the domain-invariant characteristics (\eg pose) of its input image $\x$, we employ the cycle consistency loss~\cite{choi2018stargan,kim2017discogan,zhu2017cyclegan}

\begin{equation}
\mathcal{L}_{cyc} = \mathbb{E}_{\x, \y, \yt, \z} \left[ {||\x - G(G(\x, \st), \sh)||}_{1} \right],
\end{equation}

\noindent where $\sh = {\E}_{\y}(\x)$ is the estimated style code of the input image $\x$, and $\y$ is the original domain of $\x$. By encouraging the generator $G$ to reconstruct the input image $\x$ with the estimated style code $\sh$, $\G$ learns to preserve the original characteristics of $\x$ while changing its style faithfully.

\medskip

\noindent \textbf{Full objective.} Our full objective functions can be summarized as follows: 
\begin{equation}
\begin{split}
\min_{\G, \F, \E} \max_\D \quad & \mathcal{L}_{adv} + {\lambda}_{sty} \thinspace \mathcal{L}_{sty} \\ & - {\lambda}_{ds}\thinspace\mathcal{L}_{ds} + {\lambda}_{cyc}\thinspace\mathcal{L}_{cyc},
\end{split}
\end{equation}
\noindent where ${\lambda}_{sty}$, ${\lambda}_{ds}$, and ${\lambda}_{cyc}$ are hyperparameters for each term. We also further train our model in the same manner as the above objective, using reference images instead of latent vectors when generating style codes. We provide the training details in
\ifarxiv
Appendix~\ref{sec:supple_training_details}.
\fi

%% file: sections/3.experiments.tex
\section{Experiments}
In this section, we describe evaluation setups and conduct a set of experiments. We analyze the individual components of \stargan~(\Sref{sec:components}) and compare our model with three leading baselines on diverse image synthesis (\Sref{sec:comparison}). All experiments are conducted using unseen images during the training phase. 
 
\medskip

\noindent \textbf{Baselines.} We use MUNIT \cite{huang2018munit}, DRIT \cite{lee2018drit}, and MSGAN \cite{mao2019msgan} as our baselines, all of which learn multi-modal mappings between two domains. For multi-domain comparisons, we train these models multiple times for every pair of image domains. We also compare our method with StarGAN \cite{choi2018stargan}, which learns mappings among multiple domains using a single generator. All the baselines are trained using the implementations provided by the authors.

\medskip

\noindent \textbf{Datasets.} We evaluate \stargan~on CelebA-HQ \cite{karras2018progressivegan} and our new AFHQ dataset 
\ifarxiv
(Appendix~\ref{sec:AFHQ}).
\fi
We separate CelebA-HQ into two domains of male and female, and AFHQ into three domains of cat, dog, and wildlife. Other than the domain labels, we do not use any additional information (\eg facial attributes of CelebA-HQ or breeds of AFHQ) and let the models learn such information as styles without supervision. For a fair comparison, all images are resized to $256 \times 256$ resolution for training, which is the highest resolution used in the baselines.

\medskip

\noindent \textbf{Evaluation metrics.} 
We evaluate both the visual quality and the diversity of generated images using Frech\'{e}t inception distance (FID) \cite{heusel2017fid} and learned perceptual image patch similarity (LPIPS) \cite{zhang2018lpips}. We compute FID and LPIPS for every pair of image domains within a dataset and report their average values. The details on evaluation metrics and protocols are further described in
\ifarxiv
Appendix~\ref{sec:supple_evaluation}. 
\fi

\tabAblation
\figablation

\figrefceleba
\figrandom

\subsection{Analysis of individual components}
\label{sec:components}
We evaluate individual components that are added to our baseline StarGAN using CelebA-HQ. \Tref{tab:ablation} gives FID and LPIPS for several configurations, where each component is cumulatively added on top of StarGAN.  An input image and the corresponding generated images of each configuration are shown in \Fref{fig:ablation}. The baseline configuration (\arch{a}) corresponds to the basic setup of StarGAN, which employs WGAN-GP~\cite{gulrajani2017wgangp}, ACGAN discriminator~\cite{odena2017acgan}, and depth-wise concatenation~\cite{mirza2014cgan} for providing the target domain information to the generator. As shown in \Fref{fig:ablation}a, the original StarGAN produces only a local change by applying makeup on the input image.

\smallskip

We first improve our baseline by replacing the ACGAN discriminator with a multi-task discriminator \cite{mescheder2018r1reg,liu2019funit}, allowing the generator to transform the global structure of an input image as shown in configuration (\arch{b}). Exploiting the recent advances in GANs, we further enhance the training stability and construct a new baseline (\arch{c}) by applying $R_1$ regularization~\cite{mescheder2018r1reg} and switching the depth-wise concatenation to adaptive instance normalization (AdaIN)~\cite{dumoulin2018featuretransform,huang2017adain}. Note that we do not report LPIPS of these variations in \Tref{tab:ablation}, since they are yet to be designed to produce multiple outputs for a given input image and a target domain. 

\smallskip

To induce diversity, one can think of directly giving a latent code $\z$ into the generator $G$ and impose the latent reconstruction loss ${||\z - {E}(G(\x,\z, \y))||}_{1}$~\cite{huang2018munit,zhu2017bicyclegan}. However, in a \textit{multi-domain} scenario, we observe that this baseline (\arch{d}) does not encourage the network to learn meaningful styles and fails to provide as much diversity as we expect. We conjecture that this is because latent codes have no capability in separating domains, and thus the latent reconstruction loss models \textit{domain-shared} styles (\eg color) rather than \textit{domain-specific} ones (\eg hairstyle). Note that the FID gap between baseline (\arch{c}) and (\arch{d}) is simply due to the difference in the number of output samples. 

Instead of giving a latent code into $\G$ directly, to learn meaningful styles, we transform a latent code $\z$ into a \textit{domain-specific} style code $\s$ through our proposed mapping network (\Fref{fig:networks}b) and inject the style code into the generator (\arch{e}). Here, we also introduce the style reconstruction loss (Eq.~\eqref{eqn:sty_rec_loss}). Note that each output branch of our mapping network is responsible to a particular domain, thus style codes have no ambiguity in separating domains. Unlike the latent reconstruction loss, the style reconstruction loss allows the generator to produce diverse images reflecting \textit{domain-specific} styles. Finally, we further improve the network to produce diverse outputs by adopting the diversity regularization (Eq.~\eqref{eqn:ds_loss}), and this configuration (\arch{f}) corresponds to our proposed method, \stargan. \Fref{fig:refceleba} shows that \stargan{} can synthesize images that reflect diverse styles of references including hairstyle, makeup, and beard, without hurting the source characteristics.

\figref

\medskip

\subsection{Comparison on diverse image synthesis}
\label{sec:comparison} 
In this section, we evaluate \stargan~on diverse image synthesis from two perspectives: latent-guided synthesis and reference-guided synthesis.

\smallskip

\noindent\textbf{Latent-guided synthesis.} \Fref{fig:latentguided} provides a qualitative comparison of the competing methods. Each method produces multiple outputs using random noise. For CelebA-HQ, we observe that our method synthesizes images with a higher visual quality compared to the baseline models. In addition, our method is the only model that can successfully change the entire hair styles of the source images, which requires non-trivial effort (\eg generating ears). For AFHQ, which has relatively large variations, the performance of the baselines is considerably  degraded, while our method still produces images with high quality and diverse styles.

\tabRandom
\smallskip

As shown in \Tref{tab:random}, our method outperforms all the baselines by a large margin in terms of visual quality. For both CelebA-HQ and AFHQ, our method achieves FIDs of 13.7 and 16.2, respectively, which are more than two times improvement over the previous leading method. Our LPIPS is also the highest in CelebA-HQ, which implies our model produces the most diverse results given a single input. We conjecture that the high LPIPS values of the baseline models in AFHQ are due to their spurious artifacts.

\smallskip

\noindent\textbf{Reference-guided synthesis.} To obtain the style code from a reference image, we sample test images from a target domain and feed them to the encoder network of each method. For CelebA-HQ (\Fref{fig:refcompare}a), our method successfully renders distinctive styles (\eg bangs, beard, makeup, and hair-style), while the others mostly match the color distribution of reference images. For the more challenging AFHQ (\Fref{fig:refcompare}b), the baseline models suffer from a large domain shift. They hardly reflect the style of each reference image and only match the domain. In contrast, our model renders distinctive styles (\eg breeds) of each reference image as well as its fur pattern and eye color. Note that \stargan~produces high quality images across all domains and these results are from a single generator. Since the other baselines are trained individually for each pair of domains, the output quality fluctuates across domains. For example, in AFHQ (\Fref{fig:refcompare}b), the baseline models work reasonably well in dog-to-wildlife (2nd row) while they fail in cat-to-dog (1st row).

\smallskip

\tabReference
\Tref{tab:reference} shows FID and LPIPS of each method for reference guided synthesis. For both datasets, our method achieves FID of 23.8, and 19.8, which are about 1.5$\times$ and 3.5$\times$ better than the previous leading method, respectively. The LPIPS of \stargan~is also the highest among the competitors, which implies that our model produces the most diverse results considering the styles of reference images. Here, MUNIT and DRIT suffer from mode-collapse in AFHQ, which results in lower LPIPS and higher FID than other methods. 

\smallskip

\noindent\textbf{Human evaluation.} We use the Amazon Mechanical Turk (AMT) to compare the user preferences of our method with baseline approaches. Given a pair of source and reference images, the AMT workers are instructed to select one among four image candidates from the methods, whose order is randomly shuffled. We ask separately which model offers the best image quality and which model best stylizes the input image considering the reference image. 
For each comparison, we randomly generate 100 questions, and each question is answered by 10 workers. We also ask each worker a few simple questions to detect unworthy workers. The number of total valid workers is 76. As shown in \Tref{tab:userstudy}, our method obtains the majority of votes in all instances, especially in the challenging AFHQ dataset and the question about style reflection. These results show that \stargan{} better extracts and renders the styles onto the input image than the other baselines. 
\tabUser

%% file: sections/4.discussion.tex
\section{Discussion}
\label{sec:discussion}
We discuss several reasons why \stargan{} can successfully synthesize images of diverse styles over multiple domains. First, our style code is separately generated per domain by the multi-head mapping network and style encoder. By doing so, our generator can only focus on using the style code, whose domain-specific information is already taken care of by the mapping network (\Sref{sec:components}). Second, following the insight of StyleGAN \cite{karras2019stylegan}, our style space is produced by learned transformations. This provides more flexibility to our model than the baselines~\cite{huang2018munit,lee2018drit,mao2019msgan}, which assume that the style space is a fixed Gaussian distribution (\Sref{sec:comparison}). Last but not least, our modules benefit from fully exploiting training data from multiple domains. By design, the shared part of each module should learn domain-invariant features which induces the regularization effect, encouraging better generalization to unseen samples. To show that our model generalizes over the unseen images, we test a few samples from FFHQ~\cite{karras2019stylegan} with our model trained on CelebA-HQ (\Fref{fig:ffhq}). Here, \stargan{} successfully 
captures styles of references and renders these styles correctly to the source images.

%% file: sections/5.related_work.tex
\figffhq

\section{Related work}
Generative adversarial networks (GANs)~\cite{goodfellow2014gan} have shown impressive results in many computer vision tasks such as image synthesis~\cite{brock2018biggan,lucic2019s3gan,donahue2019bigbigan}, colorization~\cite{kim2019tag2pix,yoo2019coloring} and super-resolution~\cite{ledig2017srgan,wang2018esrgan}. Along with improving the visual quality of generated images, their diversity also has been considered as an important objective which has been tackled by either devoted loss functions~\cite{mao2019msgan,mescheder2018r1reg} or architectural design~\cite{brock2018biggan,karras2019stylegan}. StyleGAN~\cite{karras2019stylegan} introduces a non-linear mapping function that embeds an input latent code into an intermediate style space to better represent the factors of variation. However, this method requires non-trivial effort when transforming a real image, since its generator is not designed to take an image as input. 

\medskip

Early image-to-image translation methods~\cite{isola2017pix2pix,zhu2017cyclegan,liu2017unit} are well known to learn a deterministic mapping even with stochastic noise inputs. Several methods reinforce the connection between stochastic noise and the generated image for diversity, by marginal matching~\cite{almahairi2018augcyclegan}, latent regression~\cite{zhu2017bicyclegan,huang2018munit}, and diversity regularization~\cite{yang2019dsgan, mao2019msgan}. Other approaches produce various outputs with the guidance of reference images~\cite{chang2018pairedcyclegan,cho2019gdwct,ma2018exemplar,park2019spade}. However, all theses methods consider only two domains, and their extension to multiple domains is non-trivial. Recently, FUNIT~\cite{liu2019funit} tackles multi-domain image translation using a few reference images from a target domain, but it requires fine-grained class labels and can not generate images with random noise. Our method provides both latent-guided and reference-guided synthesis and can be trained with coarsely labeled dataset. In parallel work, Yu et al.~\cite{liu2019multimapping} tackle the same issue but they define the style as domain-shared characteristics rather than domain-specific ones, which limits the output diversity. 

%% file: sections/6.conclusion.tex
\vspace{0mm}
\section{Conclusion}

\ifarxiv

We proposed \stargan{}, which addresses two major challenges in image-to-image translation; translating an image of one domain to diverse images of a target domain, and supporting multiple target domains. The experimental results showed that our model can generate images with rich styles across multiple domains, remarkably outperforming the previous leading methods~\cite{huang2018munit,lee2018drit, mao2019msgan}. We also released a new dataset of animal faces (AFHQ) for evaluating methods in a large inter- and intra domain variation setting.

\noindent\textbf{Acknowledgements.} We thank the full-time and visiting Clova AI members for an early review: Seongjoon Oh, Junsuk Choe, Muhammad Ferjad Naeem, and Kyungjune Baek. All experiments were conducted based on NAVER Smart Machine Learning (NSML)~\cite{kim2018nsml, sung2017nsml}.

\fi

%% file: sections/7.appendix.tex
\appendix

\ifarxiv
\figafhqdataset
\fi

\section{The AFHQ dataset}
\label{sec:AFHQ}
We release a new dataset of animal faces, Animal Faces-HQ (AFHQ), consisting of 15,000 high-quality images at $512 \times 512$ resolution. \Fref{fig:afhq_dataset} shows example images of the AFHQ dataset. The dataset includes three domains of cat, dog, and wildlife, each providing 5000 images. By having multiple (three) domains and diverse images of various breeds ($\geq$ eight) per each domain, AFHQ sets a more challenging image-to-image translation problem. For each domain, we select 500 images as a test set and provide all remaining images as a training set. We collected images with permissive licenses from the Flickr\footnote{https://www.flickr.com} and Pixabay\footnote{https://www.pixabay.com} websites. All images are vertically and horizontally aligned to have the eyes at the center. The low-quality images were discarded by human effort. 
\ifarxiv
We have made dataset available at~\url{https://github.com/clovaai/stargan-v2}.
\fi

\section{Training details} 
\label{sec:supple_training_details}
For fast training, the batch size is set to eight and the model is trained for 100K iterations. The training time is about three days on a single Tesla V100 GPU with our implementation in PyTorch~\cite{adam2017pytorch}. We set ${\lambda}_{sty}=1$, ${\lambda}_{ds}=1$,  and ${\lambda}_{cyc}=1$ for CelebA-HQ and ${\lambda}_{sty}=1$, ${\lambda}_{ds}=2$, and ${\lambda}_{cyc}=1$ for AFHQ. To stabilize the training, the weight ${\lambda}_{ds}$ is linearly decayed to zero over the 100K iterations. We adopt the non-saturating adversarial loss~\cite{goodfellow2014gan} with ${R}_{1}$ regularization~\cite{mescheder2018r1reg} using $\gamma=1$. We use the Adam~\cite{kingma2014adam} optimizer with ${\beta}_{1}=0$ and ${\beta}_{2}=0.99$. The learning rates for $\G, \D,$ and $\E$ are set to ${10}^{-4}$, while that of $F$ is set to ${10}^{-6}$. For evaluation, we employ exponential moving averages over parameters~\cite{karras2018progressivegan,yazici2018movingaveragegan} of all modules except $\D$. We initialize the weights of all modules using He initialization~\cite{he2015heinit} and set all biases to zero, except for the biases associated with the scaling vectors of AdaIN that are set to one.

\section{Evaluation protocol} 
\label{sec:supple_evaluation}
This section provides details for the evaluation metrics and evaluation protocols used in all experiments.

\smallskip

\noindent \textbf{Frech\'{e}t inception distance (FID) \cite{heusel2017fid}} measures the discrepancy between two sets of images. We use the feature vectors from the last average pooling layer of the ImageNet-pretrained Inception-V3 \cite{szegedy2016inceptionv3}. For each test image from a source domain, we translate it into a target domain using 10 latent vectors, which are randomly sampled from the standard Gaussian distribution. We then calculate FID between the translated images and training images in the target domain. We calculate the FID values for every pair of image domains (\eg female $\rightleftarrows$ male for CelebA-HQ) and report the average value. Note that, for reference-guided synthesis, each source image is transformed using 10 reference images randomly sampled from the test set of a target domain. 

\smallskip

\noindent \textbf{Learned perceptual image patch similarity (LPIPS) \cite{zhang2018lpips}} measures the diversity of generated images using the ${L}_{1}$ distance between features extracted from the ImageNet-pretrained AlexNet \cite{alex2012alexnet}. For each test image from a source domain, we generate 10 outputs of a target domain using 10 randomly sampled latent vectors. Then, we compute the average of the pairwise distances among all outputs generated from the same input (\ie 45 pairs). Finally, we report the average of the LPIPS values over all test images. For reference-guided synthesis, each source image is transformed using 10 reference images to produce 10 outputs.

\section{Additional results}
\label{sec:supple_additional_results}
We provide additional reference-guided image synthesis results on both CelebA-HQ and AFHQ (Figure~\ref{fig:refcelebatwo} and~\ref{fig:refafhq}). In CelebA-HQ, \stargan{} synthesizes the source identity in diverse appearances reflecting the reference styles such as hairstyle, and makeup. In AFHQ, the results follow the breed and hair of the reference images preserving the pose of the source images. Interpolation results between styles can be found at ~\url{https://youtu.be/0EVh5Ki4dIY}. 

\figrefcelebatwo

\figrefafhq

\section{Network architecture} 
\label{sec:supple_network_architecture}

In this section, we provide architectural details of \stargan{}, which consists of four modules described below.

\smallskip

\noindent\textbf{Generator} \textbf{(}\Tref{tab:generator_architecture}\textbf{).} For AFHQ, our generator consists of four downsampling blocks, four intermediate blocks, and four upsampling blocks, all of which inherit pre-activation residual units~\cite{he2016resblk}. We use the instance normalization~(IN)~\cite{ulyanov2016instancenorm} and the adaptive instance normalization~(AdaIN)~\cite{huang2017adain,karras2019stylegan} for down-sampling and up-sampling blocks, respectively. A style code is injected into all AdaIN layers, providing scaling and shifting vectors through learned affine transformations. For CelebA-HQ, we increase the number of downsampling and upsampling layers by one. We also remove all shortcuts in the upsampling residual blocks and add skip connections with the adaptive wing based heatmap~\cite{wang2019adawing}.


\smallskip

\noindent\textbf{Mapping network} \textbf{(}\Tref{tab:mappingnetwork_architecture}\textbf{).} Our mapping network consists of an MLP with \texttt{K} output branches, where \texttt{K} indicates the number of domains. Four fully connected layers are shared among all domains, followed by four specific fully connected layers for each domain. We set the dimensions of the latent code, the hidden layer, and the style code to 16, 512, and 64, respectively. We sample the latent code from the standard Gaussian distribution. We do not apply the pixel normalization~\cite{karras2019stylegan} to the latent code, which has been observed not to increase model performance in our tasks. We also tried feature normalizations~\cite{ba2016layernorm,ioffe2015batchnorm}, but this degraded performance.

\smallskip

\noindent\textbf{Style encoder} \textbf{(}\Tref{tab:encoder_architecture}\textbf{).} Our style encoder consists of a CNN with \texttt{K} output branches, where \texttt{K} is the number of domains. Six pre-activation residual blocks are shared among all domains, followed by one specific fully connected layer for each domain. We do not use the global average pooling~\cite{huang2018munit} to extract fine style features of a given reference image. The output dimension “\texttt{D}” in \Tref{tab:encoder_architecture} is set to 64, which indicates the dimension of the style code.

\smallskip

\noindent\textbf{Discriminator} \textbf{(}\Tref{tab:encoder_architecture}\textbf{).} Our discriminator is a multi-task discriminator~\cite{mescheder2018r1reg}, which contains multiple linear output branches~\footnote{The original implementation of the multi-task discriminator can be found at~\url{https://github.com/LMescheder/GAN\_stability}.}. The discriminator contains six pre-activation residual blocks with leaky ReLU~\cite{maas2013lrelu}. We use \texttt{K} fully-connected layers for real/fake classification of each domain, where \texttt{K} indicates the number of domains. The output dimension “\texttt{D}” is set to 1 for real/fake classification. We do not use any feature normalization techniques~\cite{ioffe2015batchnorm,ulyanov2016instancenorm} nor PatchGAN~\cite{isola2017pix2pix} as they have been observed not to improve output quality. We have observed that in our settings, the multi-task discriminator provides better results than other types of conditional discriminators~\cite{mirza2014cgan,miyato2018projectiondisc,odena2017acgan,reed2016cgan_hcat}.

\networkG
\networkF
\networkE